\documentclass{article}
\usepackage{spconf,amsmath,graphicx}
\usepackage{amssymb}
\usepackage{booktabs}
\usepackage{float}

\title{Intra-modal constraint loss for image-text retrieval}
\name{{Jianan Chen}$^{\star}$\qquad{Lu Zhang}$^{\star}$\qquad{Qiong Wang}$^{\star}$\qquad{Cong Bai}$^{\dagger}$\qquad{Kidiyo Kpalma}$^{\star}$}
  
\address{$^{\star}$Univ Rennes, INSA Rennes, CNRS, IETR-UMR 6164, F-35000 Rennes, France\\
    $^{\dagger}$College of Computer Science, Zhejiang University of Technology, Hangzhou, China} 

\begin{document}
%
\maketitle
\begin{abstract}
Cross-modal retrieval has drawn much attention in both computer vision and natural 
language processing domains. 
With the development of convolutional and recurrent neural networks, 
the bottleneck of retrieval across image-text modalities is no longer 
the extraction of image and text features but an efficient loss function learning in embedding space.
Many loss functions try to closer pairwise features from heterogeneous modalities.
This paper proposes a method for learning joint embedding of images and 
texts using an intra-modal constraint loss function to reduce the violation of 
negative pairs from the same homogeneous modality.
Experimental results show that our approach outperforms state-of-the-art 
bi-directional image-text retrieval methods on Flickr30K and Microsoft COCO datasets. 
Our code is publicly available\footnote{https://github.com/CanonChen/IMC}.
\end{abstract}
\section{Introduction}
\label{sec:intro}


In recent years, cross-modal retrieval has attracted a lot of attention in both computer 
vision and natural language processing domains. 
Image-text retrieval is a task that aims to find the most relative semantic pairs in heterogeneous modalities.
It can generally be seen as two-direction queries: one is the query by image feature to retrieve the text information by the relevant rank and vice versa.
With advances in deep neural network technology, the bottleneck in image-text retrieval has shifted from different modalities features extraction to embedding representational loss function learning.

Many loss functions have been proposed in the text-image retrieval domain.
Most recent approaches used a hinge-based triplet ranking loss~\cite{faghri2018vse++,AndrejKarpathy2015DeepVA}, also referred to as Sum of Hinges (SH) loss, to reduce the retrieval distance in both directions. Faghri et al. further proposed a Max of Hinges (MH) loss based on the SH loss, by emphasizing hard negatives for training, which achieves better performances than the SH loss~\cite{faghri2018vse++}. The existing loss functions deal well with heterogeneous modalities pairs. However, few losses consider the effect of homogeneous modality pair distances. To address this issue, we propose a novel Intra-Modal Constraint (IMC) loss to reduce the violation between negative pairs within the same modality.

In our previous survey work~\cite{chen9175503}, the deep learning based image-text 
retrieval architectures are divided into four categories: 1) ``pairwise embeddings learning'' which uses one branch to generate the image features and another one to generate the text image features, e.g.,~\cite{zhangDeepCrossModalProjection2018,faghri2018vse++}; 2) ``adversarial learning'' which introduces the Generative Adversarial Nets (GAN~\cite{goodfellow2014generative}) in the latent space; 3) ``interaction learning'' which has information transfer between the image and text branch; 4)~``attribute learning'' which involves high-level semantic intrinsic attributes through attention mechanisms or graph neural networks, etc.
Because of the conventionality and simplicity of the ``pairwise embeddings learning'', we choose to validate our proposed IMC loss on this type of architecture. 

Experimental results on two commonly used image-text retrieval datasets show that a typical ``pairwise embeddings learning'' architecture combined with our IMC loss can achieve higher performance, compared to state-of-the-art methods. The ablation study also shows the improvements brought by the IMC loss, compared to the MH loss.

Our main contributions are as follows: 

1. We propose an Intra-Modal Constraint loss to reduce the violation of negative pairs in the homogeneous modality.

2. We develop a ``pairwise embeddings learning'' network combined with our loss function for image-text retrieval.

3. We test our methods on the two most popular image-text retrieval datasets and evaluate the influence of different similarity distances in the IMC loss.

\section{The Proposed Method}
In this section, we first elaborate on the architecture in section~\ref{ssec:architecture}. 
Then we introduce our designed loss function for image-text retrieval in section~\ref{ssec:IMCL}.

\subsection{Architecture}\label{ssec:architecture}
\begin{figure}[htb]
  \centerline{\includegraphics[width=8.6cm]{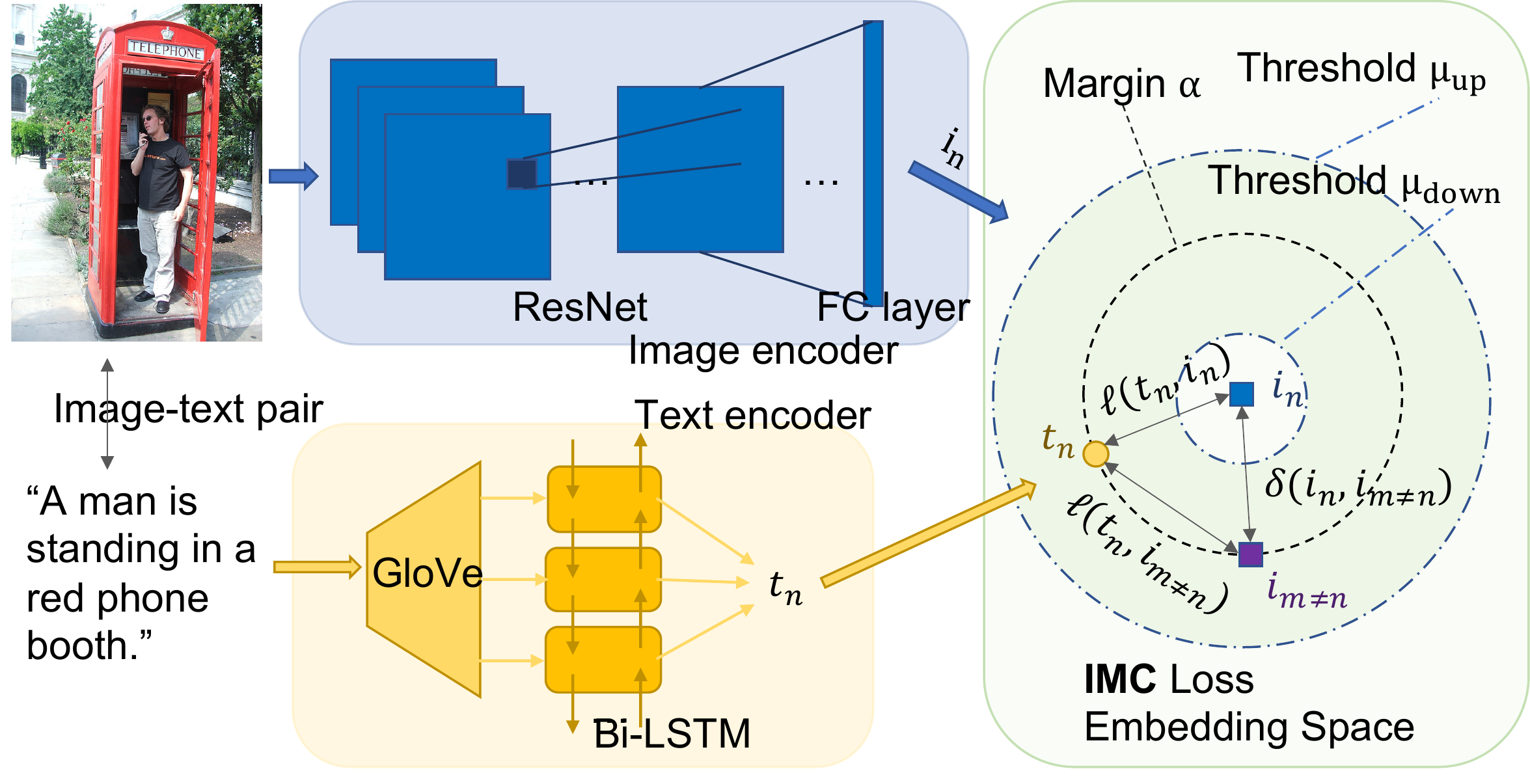}}
  \caption{The framework of Intra-Modal Constraint network. 
  Two-branch network encoders extract the image and text features on the left side. 
  The embedding space for features projection is on the right side: the intra-modal constraint loss learning reduces the distances of inter-modal pairwise 
  feature representations (i.e., yellow circle and blue square) 
  and increases the intra-modal non-pair distance (i.e., the blue square and purple square) simultaneously. 
  More details will be explained in section~\ref{ssec:IMCL}.}
  \label{fig:FW}
\end{figure}

We use a two-branch structure network to extract the image and text features, 
then project heterogeneous modalities features into a common embedding space
and learn the representation by intra-modal constraint loss, as shown in Fig.~\ref{fig:FW}, i.e., a typical ``pairwise embeddings learning'' architecture.
Here, the image encoder is the pre-trained ResNet152~\cite{He_2016_CVPR}, the same as most previous works. Then ResNet152 is followed by an additional fully connected (FC) layer to get the image feature vector $i_n$ (where $n$ is the image number index) with the same dimension $d$ ($d=1024$ here) as the text feature vector $t_n$. 
In the text encoder, we firstly get word represents via the pre-trained GloVe~\cite{pennington2014glove};
then employ a Bi-direction Long Short-Term Memory~(Bi-LSTM)~\cite{10.1162/neco.1997.9.8.1735} to obtain the final text feature vector $t_n$.
Finally, the image feature vector $i_n$ and the text feature vector $t_n$ are embedded onto a common space.

Hereafter we refer to $(i_n,t_n)$ as a positive image-text features pair coming from the
$n$-th image and its relevant text; and refer to $(i_n,t_{m\neq n})$ as a negative pair from the $n$-th image and the $m$-th text which is non-relevant to the $n$-th image.

To train the FC layer in the image encoder and the Bi-LSTM, we need to design a loss function, the minimization of which should lead to the minimization of the distance between the positive pairs and the maximization of the distance between the negative pairs in the embedding space at the same time. In the following section, we'll firstly recall two state-of-the-art loss functions (SH loss and MH loss), and then introduce our proposed Intra-Modal Constraint (IMC) loss.





\subsection{Intra-Modal Constraint Loss}\label{ssec:IMCL}
Suppose that we have N images in the training set, the total training image and text feature vectors can then be denoted as $\mathit{I=\{i_n\}_{n=1}^N}$ and $\mathit{T=\{t_n\}_{n=1}^N}$, respectively.
The most commonly used SH loss~\cite{AndrejKarpathy2015DeepVA,liuNeighborawareApproachImagetext2019} aims to minimize the cumulative
loss over training data:
\begin{small}
  \begin{equation}\label{eq:sh}
    \begin{aligned}
    \mathit{L_{\mathrm{SH}}\left(I,T\right)}=&\sum_{n,m \in N} [\alpha-\ell(i_n, t_n)+\ell(i_n, t_{m\neq n})]_{+} +\\
    &\sum_{n,m \in N} [\alpha-\ell(t_n, i_n)+\ell(t_n, i_{m \neq n})]_{+},
    \end{aligned}
  \end{equation}
\end{small}
where $\alpha$ is a margin parameter, $[x]_+\equiv max(x,0)$ and $\ell(x,y)$ is some distance function between two vectors $x$ and $y$.

The SH loss counts the distance of every negative pair with a larger distance than that of the positive pair in the margin. 
Faghri et al.~\cite{faghri2018vse++} showed that in case multiple negatives with small violations combine to dominate this loss, local minima may be created in the SH loss. Thus they proposed the MH loss, which solves this problem by focusing on the hardest negative:
\begin{small}
\begin{equation}\label{eq:mh}
  \begin{aligned}
    \mathit{L_\mathrm{MH}\left(I,T\right)}=&\max_{n,m\in N} [\alpha+\ell(i_n, t_{m\neq n})-\ell(i_n, t_n)]_{+} +\\
    &\max_{n,m\in N} [\alpha+\ell(t_n, i_{m \neq n})-\ell(t_n, i_n)]_{+}.
  \end{aligned}
\end{equation}
\end{small}
They also demonstrated that the MH loss performs better than the SH loss empirically~\cite{faghri2018vse++} .

However, both SH and MH losses focus on heterogeneous modalities pairs but ignore the negative pairs in the homogeneous modality.
To retrieve more relevant pairs of image and text, we propose a novel loss function for reducing the intra-modal pairwise effect, named Intra-Modal Constraint (IMC) loss. Our IMC loss combines the MH loss with two Intra-Modal Constraint (IMC) terms:
\begin{small}
\begin{equation}\label{eq:imcl}
  \begin{aligned}
    \mathit{L_\mathrm{IMC}\left(I,T\right)}=&\max_{n,m\in N} [\alpha+\ell(i_n, t_{m\neq n})-\ell(i_n, t_n)]_{+} +\\
    &\max_{n,m\in N} [\alpha+\ell(t_n, i_{m \neq n})-\ell(t_n, i_n)]_{+} +\\
    &{IMC(I)} + {IMC(T)},
  \end{aligned}
\end{equation}
\end{small}
where $IMC(I$) and $IMC(T)$ are the image-modal constraint and text-modal constraint, respectively. They are defined in the same way as follows:
\begin{small}
\begin{equation}\label{eq:imc}
  \mathit{IMC\left(V\right) = \lambda\sum_{n,m \in N}
  \begin{cases}
    0,\quad\delta(v_n, v_{m\neq n})\le\mu_\mathrm{down}\\
    \delta(v_n, v_{m\neq n}),\\
    0,\quad\mu_\mathrm{up}\le\delta(v_n, v_{m\neq n})
  \end{cases}}
\end{equation}
\end{small}
where $\lambda$ is a weight parameter to balance the influence of IMC terms, $\delta(x,y)$ is a similarity function between vectors $x$ and~$y$, 
and $\mu_\mathrm{up}$ and $\mu_\mathrm{down}$ are two thresholds defining boundaries. 

On the right side of Fig.~\ref{fig:FW}, we illustrate the principle of the IMC loss, which considers both positive/negative 
and inter-/intra-modal pairs. Note that we only calculate the similarity distance within the boundaries.

\begin{table*}[htp]
  \centering
  \resizebox{\textwidth}{!}{%
  \begin{tabular}{@{}ll|ccccccc|ccccccc@{}}
  \toprule
  \textbf{}                          & \textbf{}      & \multicolumn{7}{c|}{\textbf{MSCOCO 1K test images}}      & \multicolumn{7}{c}{\textbf{MSCOCO 5K test images}}       \\
  \textbf{} &
    \textbf{} &
    \multicolumn{3}{c}{\textbf{Image-query-Text}} &
    \multicolumn{3}{c}{\textbf{Text-query-Image}} &
    \textbf{} &
    \multicolumn{3}{c}{\textbf{Image-query-Text}} &
    \multicolumn{3}{c}{\textbf{Text-query-Image}} &
    \textbf{} \\ \midrule
  \multicolumn{1}{l|}{\textbf{Method}} &
    \textbf{Encoder Backbone} &
    \textbf{R@1} &
    \textbf{R@5} &
    \textbf{R@10} &
    \textbf{R@1} &
    \textbf{R@5} &
    \textbf{R@10} &
    \textbf{R-sum} &
    \textbf{R@1} &
    \textbf{R@5} &
    \textbf{R@10} &
    \textbf{R@1} &
    \textbf{R@5} &
    \textbf{R@10} &
    \textbf{R-sum} \\ \midrule
  \multicolumn{1}{l|}{GMM-FV~\cite{kleinAssociatingNeuralWord2015}}        & VGG,~GMM+HGLMM  & 39.4          & 67.9 & 80.9 & 25.1 & 59.8 & 76.6 & 349.7 & 17.3          & 39.0 & 50.2 & 10.8 & 28.3 & 40.1 & 185.7 \\
  \multicolumn{1}{l|}{DVSA~\cite{AndrejKarpathy2015DeepVA}}                & RCNN,~Bi-RNN    & 38.4          & 69.9 & 80.5 & 27.4 & 60.2 & 74.8 & 351.2 & 16.5          & 39.2 & 52.0 & 10.7 & 29.6 & 42.2 & 190.2 \\
  \multicolumn{1}{l|}{VQA-A~\cite{linLeveragingVQA2016}}                   & VGG,~GRU        & 50.5          & 80.1 & 89.7 & 37.0 & 70.9 & 82.9 & 411.1 & 23.5          & 50.7 & 63.6 & 16.7 & 40.5 & 53.8 & 248.8 \\
  \multicolumn{1}{l|}{TOP-$k$ Ranking~\cite{zhangDeepTopRanking2020}}      & VGG,~MLP        & 47.8          & 80.7 & 87.9 & 38.1 & 77.8 & 87.1 & 419.4 & -             & -    & -    & -    & -    & -    & -     \\
  \multicolumn{1}{l|}{CMPM~\cite{zhangDeepCrossModalProjection2018}}       & ResNet,~Bi-LSTM & 56.1          & 86.3 & 92.9 & 44.6 & 78.8 & 89.0 & 447.7 & 31.1          & 60.7 & 73.9 & 22.9 & 50.2 & 63.8 & 302.6 \\
  \multicolumn{1}{l|}{NAR~\cite{liuNeighborawareApproachImagetext2019}}    & ResNet,~HGLMM   & 61.3          & 87.9 & 95.4 & 47.0 & 80.8 & 90.1 & 462.5 & -             & -    & -    & -    & -    & -    & -     \\
  \multicolumn{1}{l|}{DPC~\cite{zhengDualpathConvolutionalImageText2020}}  & ResNet,~TextCNN & \textbf{65.6} & 89.8 & 95.5 & 47.1 & 79.9 & 90.0 & 467.9 & 41.2          & 70.5 & 81.1 & 25.3 & 53.4 & 66.4 & 337.9 \\
  \multicolumn{1}{l|}{VSE++\cite{faghri2018vse++}}                         & ResNet,~GRU     & 64.6          & 90.0 & 95.7 & 52.0 & 84.3 & 92.0 & 478.6 & \textbf{41.3} & 71.1 & 81.2 & 30.3 & 59.4 & 72.3 & 355.6 \\ \midrule
  \multicolumn{1}{l|}{\textbf{IMC(ours)}} &
    ResNet,~Bi-LSTM &
    65.3 &
    \textbf{90.8} &
    \textbf{96.4} &
    \textbf{53.9} &
    \textbf{86.0} &
    \textbf{93.6} &
    \textbf{486.0} &
    41.1 &
    \textbf{71.5} &
    \textbf{81.9} &
    \textbf{30.6} &
    \textbf{61.7} &
    \textbf{74.1} &
    \textbf{360.9} \\ \bottomrule
  \end{tabular}%
  }
  \caption{Comparison results with the state-of-the-art methods on MSCOCO~\cite{lin2014microsoft} (1K and 5K) dataset.
  {R@{1, 5, 10}} of two direction queries are listed and ordered by {R-sum} of 1K test. 
  The bests are in bold. 
  We collect the state-of-the-art results from their papers. 
  '-' means the result is not provided.
  The second column gives each method's backbone networks of the image and text feature encoder.}
  \label{tab:coco}
\end{table*}

In general, the similarity function $\delta$ can be the same as $\ell$ which is normally a cosine distance function in the literature:
\begin{footnotesize}
\begin{equation}\label{eq:cos}
\delta_\mathrm{cos}\left(v_n,v_{m\neq n}\right) = 1 - \cos\left(v_n,v_{m\neq n}\right) = 1 - {\frac{\sum_{k=1}^d {v_n}_k\cdot {v_{m\neq n}}_k}{\sqrt{\sum{{{v_n}_k}^2}}\sqrt{\sum{{{v_{m\neq n}}_k}^2}}}},
\end{equation}
\end{footnotesize}
where $(v_n,v_{m\neq n})$ are negative pairs in the same modality, $d$ is the vector dimension.
The similarity function $\delta$ can of course use other similarity metrics,
e.g., Mean Squared Displacement~(MSD):
\begin{equation}\label{eq:msd}
  \delta_\mathrm{msd}\left(v_n,v_{m\neq n}\right)=\mathrm{msd}\left(v_n,v_{m\neq n}\right) = \sum_{k=1}^d({v_n}_k - {v_{m\neq n}}_k)^2,
\end{equation}
Manhattan distance~(L1):
\begin{equation}\label{eq:l1}
  \delta_\mathrm{L1}\left(v_n,v_{m\neq n}\right)=\mathrm{L1}\left(v_n,v_{m\neq n}\right) = \sum_{k=1}^d\lvert {v_n}_k - {v_{m\neq n}}_k\rvert,
\end{equation}
or Euclidean distance~(L2):
\begin{equation}\label{eq:l2}
  \delta_\mathrm{L2}\left(v_n,v_{m\neq n}\right)=\mathrm{L2}\left(v_n,v_{m\neq n}\right) = \sum_{k=1}^d\sqrt{({v_n}_k - {v_{m\neq n}}_k)^2}.
\end{equation}
The influence of these different similarity functions on the final results will be shown in section \ref{sssec:influence}.





\section{Experiments}\label{sec:exp}
In this section, we evaluate our approach on two popular public datasets,
discuss the results, and analyze the contributions of the IMC loss compared to the MH loss.

\subsection{Datasets}\label{ssec:ds}
\noindent\textbf{MSCOCO}~\cite{lin2014microsoft} consists of 128K images and each one is described by five sentences.
MSCOCO is split into 82,783 training images,
5000 validation images and 5000 test images \cite{AndrejKarpathy2015DeepVA}.
We also use the rest of 30,504 images in original validation set of 
MSCOCO as training images which gives totally 113,287 images in our training set following the previous work \cite{faghri2018vse++}.
We report the results both on 5K test images and the average over 5 folds of 1K test images. 

\noindent\textbf{Flickr30K}~\cite{young-etal-2014-image} is a standard dataset for image-text retrieval,
including 30000 images.  It is split into 29K training images, 1K validation images and 1K test images~\cite{AndrejKarpathy2015DeepVA}.

  \begin{table}[htb!]
    \centering
    \resizebox{\linewidth}{!}{%
    \begin{tabular}{@{}llllllll@{}}
    \toprule
                      & \multicolumn{7}{c}{\textbf{Flickr30K}}                                                                        \\
                    & \multicolumn{3}{c}{\textbf{Image-query-Text}} & \multicolumn{3}{c}{\textbf{Text-query-Image}} &                \\ \midrule
    \textbf{Method} & \textbf{R@1}  & \textbf{R@5}  & \textbf{R@10} & \textbf{R@1}  & \textbf{R@5}  & \textbf{R@10} & \textbf{R-sum} \\ \midrule
    DVSA~\cite{AndrejKarpathy2015DeepVA}                & 22.2          & 48.2          & 61.4          & 15.2          & 37.7          & 50.5          & 235.2          \\
    VQA-A~\cite{linLeveragingVQA2016}                   & 33.9          & 62.5          & 74.5          & 24.9          & 52.6          & 64.8          & 313.2          \\
    GMM-FV~\cite{kleinAssociatingNeuralWord2015}        & 35.0          & 62.0          & 73.8          & 25.0          & 52.7          & 66.0          & 314.5          \\
    kNN-margin~\cite{liuStrongRobustBaseline2019}	      & 36.0	        & 64.4	        & 72.5          & 26.7	        & 54.3	        & 65.7	        & 319.6          \\
    TOP-$k$ Ranking~\cite{zhangDeepTopRanking2020}      & 41.3          & 70.3          & 79.8          & 33.1          & 61.5          & 72.9          & 358.9          \\
    BSSAN~\cite{huangBiDirectionalSpatialSemanticAttention2019}	            & 44.6	        & 74.9	        & 84.3	        & 33.2	        & 62.6	        & 72.9	        & 372.5          \\
    MDM~\cite{maBidirectionalImagesentenceRetrieval2019}	              & 44.9	        & 75.4	        & 84.4	        & 34.4	        & 67.0	        & 77.7	        & 383.8          \\
    CMPM+CMPC~\cite{zhangDeepCrossModalProjection2018}  & 49.6          & 76.8          & 86.1          & 37.3          & 65.7          & 75.5          & 391.0          \\
    VSE++~\cite{faghri2018vse++}                        & 52.9          & 80.5          & 87.2          & 39.6          & 70.1          & 79.5          & 409.8          \\
    NAR~\cite{liuNeighborawareApproachImagetext2019}               & 55.1          & 80.3          & 89.6          & 39.4          & 68.8          & 79.9          & 413.1          \\
    DPC~\cite{zhengDualpathConvolutionalImageText2020}               & 55.6          & 81.9          & 89.5          & 39.1          & 69.2          & 80.9          & 416.2          \\
    SCO~\cite{huang2018learning}                        & 55.5          & 82.0          & 89.3          & 41.1          & 70.5          & 80.1          & 418.5          \\
    CVSE++~\cite{zhouLadderLossCoherent2020}            & 56.6          & 82.5          & 90.2          & 42.4          & 71.6          & 80.8          & 424.1          \\
    CMKA~\cite{chenCrossModalKnowledgeAdaptation2021}              & 55.7          & 82.9          & 90.0          & 45.0          & 73.4          & 82.7          & 429.7          \\
    \textbf{IMC(ours)}     &  \textbf{58.5}          & \textbf{85.0} & \textbf{91.2}          & \textbf{45.4} & \textbf{74.8}          & \textbf{83.0}          & \textbf{437.9} \\ \bottomrule
    \end{tabular}%
    }
    \caption{Comparison results on Flickr30K}
    \label{tab:f30k}
  \end{table}

\subsection{Settings and performance metrics}\label{ssec:details}
\noindent\textbf{Implementation details} The model is implemented in PyTorch with
a NVIDIA 2080Ti GPU. We resize and crop the input images in the same way 
as Faghti et al.~\cite{faghri2018vse++}. The Bi-LSTM is initialized with Xavier init.~\cite{glorot2010understanding} and uses
the dropout with a probability of $0.5$ to avoid overfitting. We train image and text encoders using Adam \cite{DiederikPKingma2015AdamAM}
optimizer, set the mini-batch size to 128 and the learning rate to 0.0002 with decay every 15 epochs. The model is trained for 30 epochs.
The distance function $\ell$ used in Eq.\ref{eq:imcl} is the cosine distance function.
Following~\cite{faghri2018vse++}, we set the margin $\alpha$ in Eq.\ref{eq:imcl} to 0.2 in all experiments.
The thresholds $\mu_\mathrm{down}$ and $\mu_\mathrm{up}$ in Eq.\ref{eq:imc} are empirically set to $0.05$ and $0.5$, respectively.

\noindent\textbf{Evaluation Metrics:}
We evaluate our experimental results by R@K and R-sum metrics.
R@K is the abbreviation of Recall at $K$, the proportion of correct matches in the top $K=[1,5,10]$
of retrieving rank. R-sum is defined as:
\begin{equation}
  \begin{small}
  \text{R-sum}=\overbrace{\text{R@1+R@5+R@10}}^{\text{Image-query-Text}}\text{+}\overbrace{\text{R@1+R@5+R@10}}^{\text{Text-query-Image}}.
  \end{small}
\end{equation}

\subsection{Experimental Results}\label{ssec:era}
Table~\ref{tab:coco} and Table~\ref{tab:f30k} show the results of our approach on the MSCOCO~\cite{lin2014microsoft} dataset and Flickr30K~\cite{young-etal-2014-image} dataset, respectively. Here, the used similarity function is L1 distance as defined in Eq.\ref{eq:l1}
and the parameter $\lambda$ is set to 1.

We can see that our approach (a typical ``pairwise embeddings learning'' architecture combined with IMC loss) achieves the highest performance in most cases on MSCOCO, except that DPC has the highest R@1 value on MSCOCO 1K test images and the VSE++ has the highest R@1 value on MSCOCO 5K test image for the Image-query-Text task. Our approach also achieves the highest performance in terms of all the metrics for both the Image-query-Text and the Text-query-Image tasks on Flickr30K.





  \begin{table}[htb!]
    \centering
    \resizebox{\linewidth}{!}{%
    \begin{tabular}{@{}llllllll@{}}
    \toprule
                      & \multicolumn{7}{c}{\textbf{Flickr30K}}                                                                        \\
                    & \multicolumn{3}{c}{\textbf{Image-query-Text}} & \multicolumn{3}{c}{\textbf{Text-query-Image}} &                \\ \midrule
    \textbf{IMC} & \textbf{R@1}  & \textbf{R@5}  & \textbf{R@10} & \textbf{R@1}  & \textbf{R@5}  & \textbf{R@10} & \textbf{R-sum} \\ \midrule
    $\lambda=0$ (i.e. MH loss)                      & 57.1          & 83.7          & 90.9          & 44.5          & 74.5          & 83.2          & 433.9          \\
    $\lambda=1, \delta_\mathrm{msd}$~(Eq.\ref{eq:msd})    & 56.7          & 83.5          & \textbf{91.4} & 44.9          & \textbf{75.5} & 83.2          & 435.2          \\
    $\lambda=1, \delta_{\cos}$~(Eq.\ref{eq:cos})   & 57.4          & 84.1          & 90.9          & 44.9          & 75.1          & 83.3          & 435.9          \\
    $\lambda=1, \delta_\mathrm{L2}$~(Eq.\ref{eq:l2})     & 58.0          & 84.2          & 90.5          & 44.7          & 75.1          & \textbf{83.4}          & 435.9          \\
    $\lambda=1, \delta_\mathrm{L1}$~(Eq.\ref{eq:l1})     & \textbf{58.5}          & \textbf{85.0} & 91.2          & \textbf{45.4} & 74.8          & 83.0          & \textbf{437.9} \\ \bottomrule
    \end{tabular}%
    }
    \caption{Influence of different similarity distances in IMC}
    \label{tab:as}
  \end{table}

\subsection{Ablation study}\label{ssec:as}
Table~\ref{tab:as} clearly shows the improvement brought by our IMC loss by an ablation study on the Flickr30K dataset. 
When $\lambda=0$, the $IMC(I)$ and $IMC(T)$ in Eq.\ref{eq:imcl} are equal to zero, thus it indicates that the MH loss is used. 
With $\lambda=1$, we show the results using different similarity functions ($\delta$ in Eq.\ref{eq:imc}). We can see that whatever the similarity function is used, the IMC loss achieves better results than the MH loss in general. 

From Table~\ref{tab:as}, we can also observe the influences of different similarity distance on the results and conclude that using the L1 distance ($\delta_\mathrm{L1}$) can achieve the best performances in general, especially in terms of R@1 and R-sum. 

\begin{figure}[htb!]
  \centering
  \begin{minipage}[b]{\linewidth}
    \centering
    \centerline{\includegraphics[width=8cm]{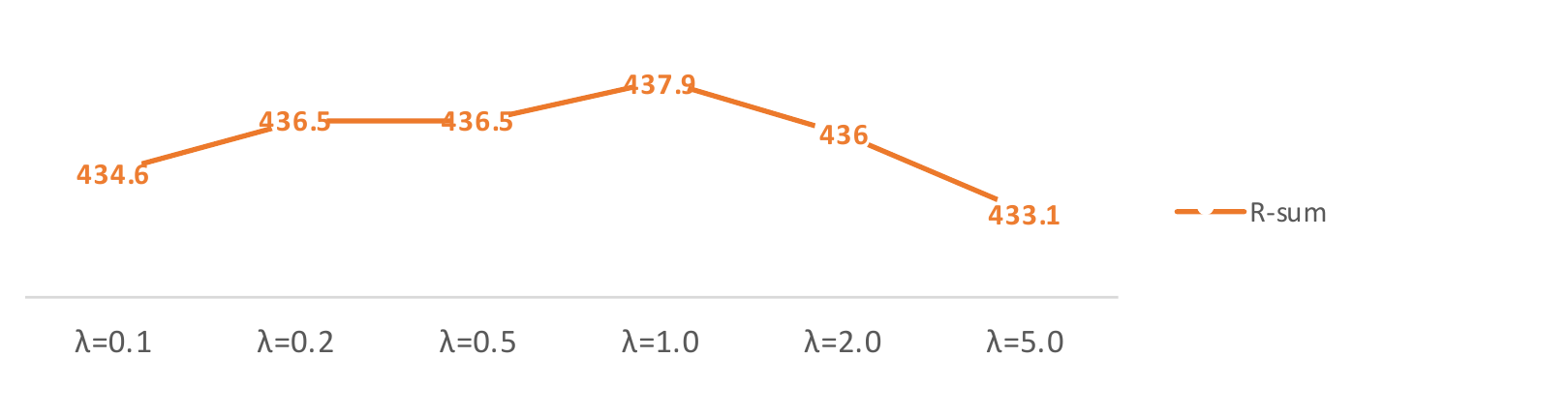}}
  \end{minipage}
  
  \begin{minipage}[b]{\linewidth}
    \centering
    \centerline{\includegraphics[width=8cm]{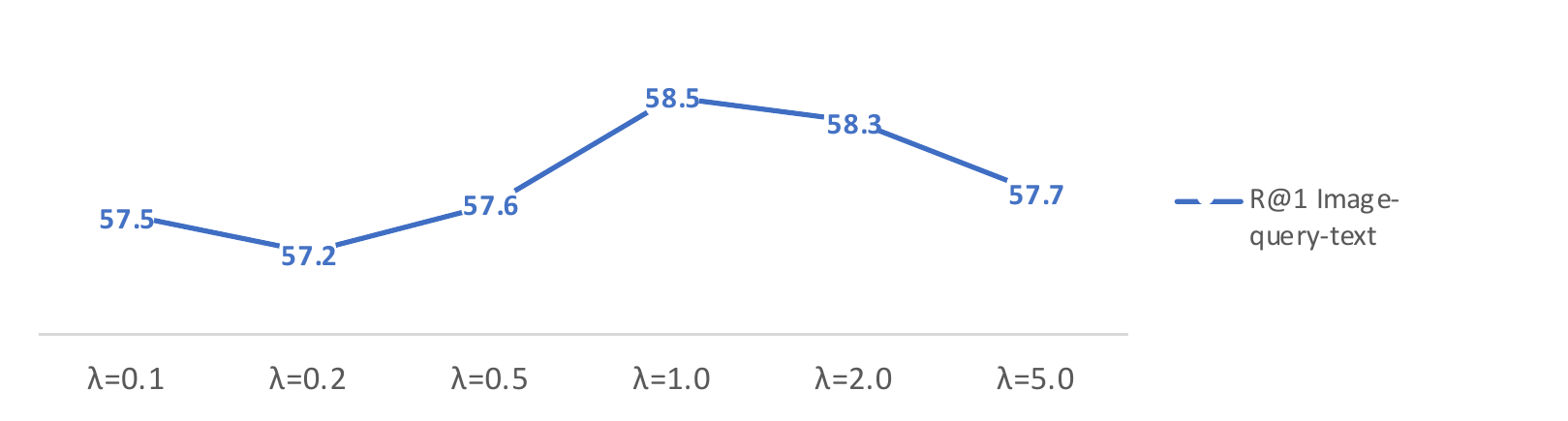}}
  \end{minipage}

  \begin{minipage}[b]{\linewidth}
    \centering
    \centerline{\includegraphics[width=8cm]{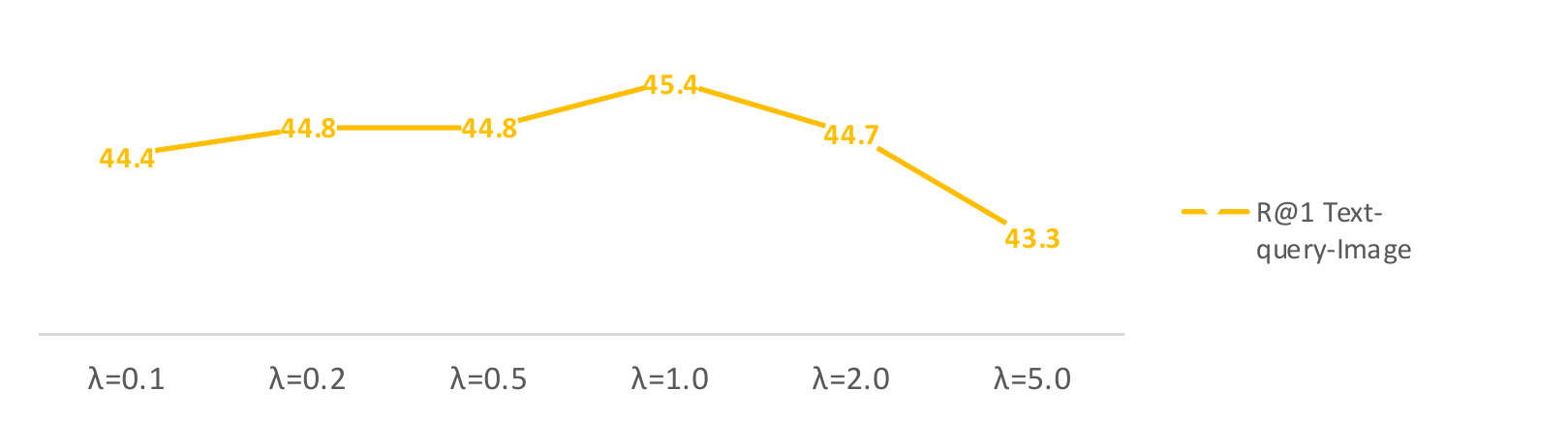}}
  \end{minipage}
  \caption{Experimental results with different values of $\lambda$ on Flickr30K.}
  \label{fig:lambda}
\end{figure}

\subsection{Influence of the weight parameter $\lambda$}\label{sssec:influence}
We vary $\lambda$ to evaluate the influence of this weight parameter in Eq.\ref{eq:imc}.
Fig.~\ref{fig:lambda} shows part of the results using $\delta_\mathrm{L1}$ on Flickr30K.
With the increase of $\lambda$, intra-modal pairs gain more emphasis.
The experimental results peak at $\lambda=1$ and then decline.
This may indicate that intra-modal and inter-modal pairs in the IMC loss are equally important.

\section{Conclusion}
In this paper, we propose a new loss (IMC loss) adapted for image-text retrieval and demonstrate its effectiveness using a two-branch ``pairwise embeddings learning'' network on two popular datasets. Our network outperforms tested state-of-the-art image-text retrieval methods and our IMC loss can improve the network's performances, compared to the MH loss. Without loss of generality, the IMC loss can also be used with other three categories of network architecture~\cite{chen9175503}. It may also improve their performances, which will be verified in our future work.




\section{Acknowledgement}
This work is supported in part by the China Scholarship Council (CSC) No.~201801810074. National Natural Science Foundation of China under Grant No.~62002326,~61976192.

\bibliographystyle{IEEEbib}
\bibliography{refs}

\begin{thebibliography}{10}

\bibitem{faghri2018vse++}
Fartash Faghri, David~J Fleet, Jamie~Ryan Kiros, and Sanja Fidler,
\newblock ``Vse++: Improving visual-semantic embeddings with hard negatives,''
\newblock in {\em Proceedings of the British Machine Vision Conference
  {(BMVC)}}, 2018.

\bibitem{AndrejKarpathy2015DeepVA}
Andrej Karpathy and Li~Fei-Fei,
\newblock ``Deep visual-semantic alignments for generating image
  descriptions,''
\newblock in {\em Computer Vision and Pattern Recognition}, 2015.

\bibitem{chen9175503}
Jianan Chen, Lu~Zhang, Cong Bai, and Kidiyo Kpalma,
\newblock ``Review of recent deep learning based methods for image-text
  retrieval,''
\newblock in {\em 2020 IEEE Conference on Multimedia Information Processing and
  Retrieval (MIPR)}, 2020, pp. 167--172.

\bibitem{zhangDeepCrossModalProjection2018}
Ying Zhang and Huchuan Lu,
\newblock ``Deep cross-modal projection learning for image-text matching,''
\newblock in {\em Proceedings of the European Conference on Computer Vision
  (ECCV)}, September 2018.

\bibitem{goodfellow2014generative}
Ian Goodfellow, Jean Pouget-Abadie, Mehdi Mirza, Bing Xu, David Warde-Farley,
  Sherjil Ozair, Aaron Courville, and Yoshua Bengio,
\newblock ``Generative adversarial nets,''
\newblock {\em Advances in neural information processing systems}, vol. 27,
  2014.

\bibitem{He_2016_CVPR}
Kaiming He, Xiangyu Zhang, Shaoqing Ren, and Jian Sun,
\newblock ``Deep residual learning for image recognition,''
\newblock in {\em Proceedings of the IEEE Conference on Computer Vision and
  Pattern Recognition (CVPR)}, June 2016.

\bibitem{pennington2014glove}
Jeffrey Pennington, Richard Socher, and Christopher~D. Manning,
\newblock ``Glove: Global vectors for word representation,''
\newblock in {\em Empirical Methods in Natural Language Processing (EMNLP)},
  2014, pp. 1532--1543.

\bibitem{10.1162/neco.1997.9.8.1735}
Sepp Hochreiter and J\"{u}rgen Schmidhuber,
\newblock ``Long short-term memory,''
\newblock {\em Neural Comput.}, vol. 9, no. 8, pp. 1735–1780, nov 1997.

\bibitem{liuNeighborawareApproachImagetext2019}
Chunxiao Liu, Zhendong Mao, Wenyu Zang, and Bin Wang,
\newblock ``A neighbor-aware approach for image-text matching,''
\newblock in {\em ICASSP 2019 - 2019 IEEE International Conference on
  Acoustics, Speech and Signal Processing (ICASSP)}, 2019, pp. 3970--3974.

\bibitem{kleinAssociatingNeuralWord2015}
Benjamin Klein, Guy Lev, Gil Sadeh, and Lior Wolf,
\newblock ``Associating neural word embeddings with deep image representations
  using fisher vectors,''
\newblock in {\em Proceedings of the IEEE Conference on Computer Vision and
  Pattern Recognition (CVPR)}, June 2015.

\bibitem{linLeveragingVQA2016}
Xiao Lin and Devi Parikh,
\newblock ``Leveraging visual question answering for image-caption ranking,''
\newblock in {\em Computer Vision -- ECCV 2016}, Bastian Leibe, Jiri Matas,
  Nicu Sebe, and Max Welling, Eds., Cham, 2016, pp. 261--277, Springer
  International Publishing.

\bibitem{zhangDeepTopRanking2020}
Lingling Zhang, Minnan Luo, Jun Liu, Xiaojun Chang, Yi~Yang, and Alexander~G.
  Hauptmann,
\newblock ``Deep top-$k$ ranking for image–sentence matching,''
\newblock {\em IEEE Transactions on Multimedia}, vol. 22, no. 3, pp. 775--785,
  2020.

\bibitem{zhengDualpathConvolutionalImageText2020}
Zhedong Zheng, Liang Zheng, Michael Garrett, Yi~Yang, Mingliang Xu, and Yi-Dong
  Shen,
\newblock ``Dual-path convolutional image-text embeddings with instance loss,''
\newblock {\em ACM Transactions on Multimedia Computing, Communications, and
  Applications (TOMM)}, vol. 16, no. 2, pp. 1--23, 2020.

\bibitem{lin2014microsoft}
Tsung-Yi Lin, Michael Maire, Serge Belongie, James Hays, Pietro Perona, Deva
  Ramanan, Piotr Doll{\'a}r, and C~Lawrence Zitnick,
\newblock ``Microsoft coco: Common objects in context,''
\newblock in {\em European conference on computer vision}. Springer, 2014, pp.
  740--755.

\bibitem{young-etal-2014-image}
Peter Young, Alice Lai, Micah Hodosh, and Julia Hockenmaier,
\newblock ``{From image descriptions to visual denotations: New similarity
  metrics for semantic inference over event descriptions},''
\newblock {\em Transactions of the Association for Computational Linguistics},
  vol. 2, pp. 67--78, 02 2014.

\bibitem{liuStrongRobustBaseline2019}
Fangyu Liu and Rongtian Ye,
\newblock ``A strong and robust baseline for text-image matching,''
\newblock in {\em Proceedings of the 57th Annual Meeting of the Association for
  Computational Linguistics: Student Research Workshop}, 2019, pp. 169--176.

\bibitem{huangBiDirectionalSpatialSemanticAttention2019}
Feiran Huang, Xiaoming Zhang, Zhonghua Zhao, and Zhoujun Li,
\newblock ``Bi-directional spatial-semantic attention networks for image-text
  matching,''
\newblock {\em IEEE Transactions on Image Processing}, vol. 28, no. 4, pp.
  2008--2020, 2019.

\bibitem{maBidirectionalImagesentenceRetrieval2019}
Lin Ma, Wenhao Jiang, Zequn Jie, and Xu~Wang,
\newblock ``Bidirectional image-sentence retrieval by local and global deep
  matching,''
\newblock {\em Neurocomputing}, vol. 345, pp. 36--44, 2019,
\newblock Deep Learning for Intelligent Sensing, Decision-Making and Control.

\bibitem{huang2018learning}
Yan Huang, Qi~Wu, Chunfeng Song, and Liang Wang,
\newblock ``Learning semantic concepts and order for image and sentence
  matching,''
\newblock in {\em Proceedings of the IEEE Conference on Computer Vision and
  Pattern Recognition}, 2018, pp. 6163--6171.

\bibitem{zhouLadderLossCoherent2020}
Mo~Zhou, Zhenxing Niu, Le~Wang, Zhanning Gao, Qilin Zhang, and Gang Hua,
\newblock ``Ladder loss for coherent visual-semantic embedding,''
\newblock {\em Proceedings of the AAAI Conference on Artificial Intelligence},
  vol. 34, no. 07, pp. 13050--13057, Apr. 2020.

\bibitem{chenCrossModalKnowledgeAdaptation2021}
Yucheng Chen, Rui Huang, Hong Chang, Chuanqi Tan, Tao Xue, and Bingpeng Ma,
\newblock ``Cross-modal knowledge adaptation for language-based person
  search,''
\newblock {\em IEEE Transactions on Image Processing}, vol. 30, pp. 4057--4069,
  2021.

\bibitem{glorot2010understanding}
Xavier Glorot and Yoshua Bengio,
\newblock ``Understanding the difficulty of training deep feedforward neural
  networks,''
\newblock in {\em Proceedings of the thirteenth international conference on
  artificial intelligence and statistics}. JMLR Workshop and Conference
  Proceedings, 2010, pp. 249--256.

\bibitem{DiederikPKingma2015AdamAM}
Diederik~P. Kingma and Jimmy Ba,
\newblock ``Adam: A method for stochastic optimization,''
\newblock in {\em International Conference on Learning Representations}, 2015.

\end{thebibliography}

\end{document}